член-корр. Б.В.Крыжановский, М.В.Крыжановский, М.Ю.Мальсагов
# ДИСКРЕТИЗАЦИЯ МАТРИЦЫ В ЗАДАЧЕ БИНАРНОЙ МИНИМИЗАЦИИ КВАДРАТИЧНОГО ФУНКЦИОНАЛА

1. В настоящей работе исследована возможность оптимальной дискретизации элементов матрицы $A_{ij}$ (замена их целочисленными значениями) в задаче минимизации квадратичного функционала

$$E = -\tfrac{1}{2}\sum_{i=1}^{N}\sum_{j=1}^{N}s_i A_{ij} s_j - \sum_{i=1}^{N} s_i B_i \qquad (1)$$

определенного в $N$-мерном конфигурационном пространстве – пространстве состояний $\mathbf{S}=(s_1,s_2,...,s_N)$ с бинарными переменными $s_i = \pm 1$ ($i=\overline{1,N}$), $\mathbf{B}\in R^N$ – вектор с вещественными компонентами $B_i$. Матрицу $A_{ij}$ без потери общности можно полагать симметричной ($A_{ij}=A_{ji}$), а ее диагональные элементы можно положить равными нулю ($A_{ii}=0$). Анализ проведем для случая, когда матричные элементы $A_{ij}$ можно рассматривать как случайные независимые величины с средним $A_0$ и стандартным отклонением $\sigma_A$.

Вычислительная сложность работы алгоритма минимизации и требования к оперативной памяти возрастают пропорционально $N^2$ и уже при относительно невысокой размерности задачи ($N \sim 10^4$) становятся трудно выполнимыми. Цель настоящей работы заключается в снижении требований к алгоритму за счет дискретизации матрицы $A_{ij}$ - выделении из $A_{ij}$ среднего значения $A_0$ и замены центрированного остатка $A'_{ij} = A_{ij} - A_0$ матрицей $a_{ij}$, нормированные элементы которой имеют целочисленные значения.

2. Алгоритм минимизации. Для решения задач бинарной оптимизации широко и весьма успешно применяются нейросетевые подходы. Наиболее часто используется модель Хопфилда [1], которую и мы примем за основу процедуры минимизации. Это полносвязная рекуррентная нейронная сеть из $N$ нейронов, имеющих два состояния $s_i = \pm 1$, $i=\overline{1,N}$. Энергия сети задана выражением (1). Такую сеть можно рассматривать как систему, решающую задачу бинарной минимизации: конвергируя в устойчивое состояние сеть находит конфигурацию, соответствующую минимуму энергии $E$. Это свойство нейронной сети успешно используется для решения различных $NP$-полных задач, решения систем линейных уравнений большой размерности и иных задач, алгоритм которых включает в себя процедуры перемножения матриц и векторов. Анализ причин подобного успеха проведен в работах [2,3], где показано, что при случайном поиске вероятность отыскания какого-либо минимума экспоненциально возрастает с ростом глубины этого минимума. Это означает, что нейросеть с подавляющей вероятностью находит если не оптимальное решение (глобальный минимум), то одно из субоптимальных решений (локальный минимум).

Мы будем рассматривать только асинхронную динамику сети Хопфилда, однозначно приводящую к минимизации функционала энергии $E$: на каждом такте работы сети вычисляется одна из компонент (например $i$-я) локального поля

$$H_i = -B_i + \sum_{j \neq i}^{N} A_{ij} s_j , \qquad (2)$$

и компоненте конфигурационного вектора $s_i$ присваивается значение $s_i = sign\, H_i$. Причем, на одном такте может быть изменено значение только одной компоненты конфигурационного вектора. Эта процедура последовательно применяется ко всем компонентам $s_i = \pm 1$ ($i = \overline{1, N}$) до тех пор, пока сеть не конвергирует в устойчивое состояние, соответствующее минимуму энергии. Описанная динамика есть не что иное, как адаптированный к бинарным переменным покомпонентный градиентный спуск ($H_i = -\partial E / \partial s_i$) по поверхности, описываемой выражением (1).

3. Процедура дискретизации. Эффективность алгоритма минимизации существенно зависит как от вида распределения матричных элементов $A_{ij}$, так и от вида процедуры дискретизации. Ниже мы будем рассматривать тип дискретизации, позволяющий не только снизить требования к оперативной памяти, но и максимально ускорить рассматриваемый алгоритм [4]. Он заключается в замене матрицы $A_{ij} = A_0 + A'_{ij}$ матрицей $a_0 + a_{ij}$, где константа $a_0$ будет определена ниже как функция $A_0$. Матрица $a_{ij}$ строится следующим образом: разобьем область распределения элементов центрированного остатка $A'_{ij} = A_{ij} - A_0$ на $(2m+1)$ отрезков; на каждом из отрезков заменим величины $A'_{ij}$ средним по отрезку значением; длины отрезков выбираем так, чтобы средние значения на отрезках были кратны величине $C$, где $C > 0$ - наименьшее положительное среднее (среднее на $k$-м отрезке равно $kC$, $k = 0, \pm 1, ..., \pm m$). В итоге таких замен получим искомую матрицу $a_{ij}$, элементы которой имеют $(2m+1)$ градаций и задаются в виде:

$$a_{ij} = kC \text{ если } (2k-1)C < A_{ij} - A_0 \leq (2k+1)C, \ k = 0, \pm 1, ..., \pm m. \qquad (3)$$

Заметим, что длины всех отрезков однозначно определяются длиной $l_0$ отрезка с нулевым средним, которая является свободным параметром для последующей оптимизации. Более того, введение отрезка с нулевым средним является ключевым моментом, позволяющим наилучшим образом представить матрицу $A_{ij} = A_0 + A'_{ij}$ матрицей $a_0 + a_{ij}$ и существенно повысить эффективность алгоритма минимизации.

Адаптируем алгоритм минимизации к работе с матрицей $a_{ij}$. Для этого сделаем в (2) замены $A_{ij} \rightarrow a_0 + a_{ij}$ и $B_i \rightarrow b_i$. Тогда, расчет локального поля и обновление компонент конфигурационного вектора будет производиться в соответствии с выражениями:

$$h_i = -b_i + \sum_{j \neq i}^{N}(a_0 + a_{ij})s_j, \quad s_i = sign\, h_i, \qquad (4)$$

где величина $a_0$ и компоненты вектора $\boldsymbol{b} = (b_1, b_2, ..., b_N)$ будут определены ниже из условия оптимизации минимизационного процесса. Заметим, что решающее правило (4) соответствует спуску по поверхности, описываемой «дискретизированным» функционалом $\varepsilon = -\tfrac{1}{2}\boldsymbol{S}(a_0 + \hat{a})\boldsymbol{S}^+ - \boldsymbol{b}\boldsymbol{S}^+$, являющимся функцией Ляпунова для алгоритма минимизации.

Заменить правило обновления $s_i = sign\, H_i$ его дискретизированным аналогом $s_i = sign\, h_i$ можно только в том случае, когда знаки локальных полей $H_i$ и $h_i$ совпадают в любой точке $N$-мерного пространства с достаточно большей вероятностью (амплитуды полей $H_i$ и $h_i$ не играют роли). Иными словами, использовать модифицированное правило градиентного спуска (4) для минимизации функционала (1) можно, если свести до минимума величину ошибки – вероятность несовпадения направлений локальных полей:

$$P = \mathbf{Pr}\{H_i h_i < 0\}. \qquad (5)$$

Для минимизации ошибки надо найти наилучшее представление матрицы $A'_{ij}$ матрицей $a_{ij}$ и найти оптимальные значения величин $a_0$ и $b_i$ ($i = \overline{1, N}$). Общее выражение для величины ошибки мы не приводим ввиду его громоздкости. Анализ этого выражения показывает, что вероятность $P$ максимальна в случайной точке пространства, с которой начинается конвергенция сети, и убывает по мере спуска в минимум. Поэтому ниже мы приведем только два предельных выражения для ошибки: в случайной точке пространства и в минимуме. На примере анализа величины $P$ в случайной точке пространства покажем как следует выбирать величины $a_0$ и $b_i$ ($i = \overline{1, N}$), как зависит ошибка от параметра дискретизации $m$ и к чему сводится наилучшее представление матрицы $A'_{ij}$ матрицей $a_{ij}$.

4. **Случайная точка.** Рассмотрим величину ошибки $P$ в случайной точке пространства. В соответствии с центральной предельной теоремой, в случае $N \gg 1$ величины $H_i$ и $h_i$ можно рассматривать как случайные (не независимые) переменные, подчиняющиеся двумерному нормальному распределению. Распределение задается средними значениями $\langle H \rangle = B_i$ и $\langle h \rangle = b_i$ величин $H_i$ и $h_i$ соответственно, их стандартными отклонениями $\sigma_H$ и $\sigma_h$ и коэффициентом корреляции $\rho$:

$$\sigma_H^2 = N(A_0^2 + \sigma_A^2), \qquad \sigma_h^2 = N(a_0^2 + \sigma_a^2), \qquad \rho = \frac{N[a_0 A_0 + \langle a_{ij} A'_{ij} \rangle]}{\sigma_H \sigma_h}, \qquad (6)$$

где $\sigma_A$ и $\sigma_a$ - стандарты матричных элементов $A_{ij}$ и $a_{ij}$ соответственно, а угловые скобки означают усреднение по ансамблю.

С учетом (6) выражение (5) преобразуется к виду:

$$P = 1 - \frac{1}{2\pi \sigma_H \sigma_h \sqrt{1-\rho^2}} \int_0^\infty \int_0^\infty (f_+ + f_-) dH dh, \qquad (7)$$

где

$$f_\pm = \exp\left\{-\frac{1}{2(1-\rho^2)}\left[\left(\frac{H \pm \langle H \rangle}{\sigma_H}\right)^2 - 2\rho\left(\frac{H \pm \langle H \rangle}{\sigma_H}\right)\left(\frac{h \pm \langle h \rangle}{\sigma_h}\right) + \left(\frac{h \pm \langle h \rangle}{\sigma_h}\right)^2\right]\right\}. \qquad (8)$$

Теперь определимся с выбором величин $a_0$ и $b_i$ ($i = \overline{1.N}$). Зададим их минимизируя ошибку $P$. Из условий $\partial P / \partial b_i = 0$ и $\partial P / \partial a_0 = 0$ найдем оптимальные значения $a_0$ и $b_i$ и соответствующее им значение коэффициента корреляции:

$$a_0 = A_0 \frac{\sigma_a^2}{\langle a_{ij} A'_{ij} \rangle}, \qquad b_i = B_i \frac{\sigma_a^2}{\langle a_{ij} A'_{ij} \rangle}, \qquad \rho = \sqrt{\frac{A_0^2 + \langle a_{ij} A'_{ij} \rangle^2 \sigma_a^{-2}}{A_0^2 + \sigma_A^2}}. \qquad (9)$$

Анализ (7)-(9) показывает, что ошибка $P$ не зависит от знаков величин $A_0$ и $B_i$. С ростом величин $|A_0|$ и $|B_i|$ величина ошибки $P$ быстро уменьшается (рис.1). Наихудшие условия для дискретизации имеют место в случае, когда $A_0 = 0$ и $B_i = 0$. В этом случае величина ошибки максимальна и описывается вытекающим из (7) выражением:

$$P = \frac{1}{2} - \frac{1}{\pi} \arcsin \rho_{\min}, \qquad (10)$$

где $\rho_{\min}$ - минимальное значение коэффициента корреляции, имеющее место при $A_0 = 0$:

$$\rho_{\min} = \frac{\langle a_{ij} A'_{ij} \rangle}{\sigma_a \sigma_A}. \qquad (11)$$

Отметим, что выражения (6)-(11) получены в самом общем виде, без привязки к виду распределения матричных элементов $A_{ij}$ и типу выбранного метода дискретизации.

Поэтому можно сделать общий вывод: наилучшее представление матрицы $A'_{ij}$ матрицей $a_{ij}$ сводится к такому подбору метода дискретизации, при котором величина $\rho_{\min}$ максимальна. В этом смысле выбранный нами метод дискретизации, направленный на максимальное ускорение алгоритма минимизации, как правило, не является наилучшим.

Выражениями (10)-(11) задается минимальное значение эффективности алгоритма минимизации. Эти выражения следует оптимизировать по длине отрезка с нулевым средним, находя $l_0$ из условия $\partial P / \partial l_0 = 0$. Для получения численных оценок и простоты анализа примем, что величины $A'_{ij}$ равномерно распределены на отрезке [-1,+1]. Тогда, из условия $\partial P / \partial l_0 = 0$ следует, что дискретизация матрицы сводится к разбиению отрезка [-1,+1] на $(2m+1)$ отрезков равной длины. В этом случае, параметры дискретизации и величина ошибки $P$ задаются выражениями:

$$C = \frac{2}{2m+1}, \quad \langle a_{ij} A'_{ij} \rangle = \sigma_a^2 = \sigma_A^2 \left[ 1 - \frac{1}{(2m+1)^2} \right],$$

$$a_0 = A_0, \quad b_i = B_i, \quad \rho = \sqrt{1 - \frac{\sigma_A^2}{(2m+1)^2 (A_0^2 + \sigma_A^2)}}. \tag{12}$$

Как видим, величина коэффициента корреляции близка к единице. Соответственно близка к нулю величина ошибки. Сказанное справедливо даже в наихудшей ситуации ($A_0 = 0$, $B_i = 0$), когда коэффициент корреляции минимален, а величина ошибки максимальна. В этом случае из (10)-(12) следуют простые оценочные выражения:

$$\rho_{\min} \approx 1 - \frac{1}{2(2m+1)}, \quad P_{\max} \approx \frac{1}{\pi(2m+1)}. \tag{13}$$

Как видим, даже в простейшем случае $m=1$, величина ошибки достаточно мала ($P_{\max} \approx 0.11$). С ростом числа градаций величина $P_{\max}$ быстро стремится к нулю. Сказанное подкрепляется результатами численного эксперимента. На рисунке 1 приведена зависимость величины ошибки от величины $|A_0|$ при различных значениях $|B_i|$. Графики приведены для случая $m=1$, однако дают хорошее представление и для произвольного значения $m$: в соответствии с (13) уменьшив в $m$ раз масштаб всех кривых на рис.1 получим графики, хорошо совпадающие как с теорией, так и данными эксперимента.

5. Минимум функционала. Процесс минимизации функционала (1) начинается с некоторой случайной точки пространства $\mathbf{S}$. Подчиняясь решающему правилу (4) сеть конвергирует в некоторое устойчивое состояние $\mathbf{S}_0^*$, являющееся минимумом

«дискретизированного» функционала $\varepsilon$. Если из этой точки продолжить спуск с решающим правилом (2), то сеть конвергирует в состояние $\mathbf{S_0}$, соответствующее минимуму функционала (1). На всем пути $\mathbf{S} \to \mathbf{S_0^*} \to \mathbf{S_0}$ величина ошибки только уменьшается. Величину ошибки в исходной (случайной) точке мы анализировали выше (7). Здесь мы рассмотрим ее величину $P = \Pr\{H_i h_i < 0 / H_i s_{0i} > 0\}$ в конечной точке $\mathbf{S_0} = (s_{01}, s_{02}, ..., s_{0N})$. Ограничимся анализом ситуации $A_0 = 0$, $B_i = 0$ ($i = \overline{1.N}$), наихудшей с точки зрения эффективности минимизации. Тогда, используя результаты работ [5,6] и проводя аналогичные (4)-(6) промежуточные выкладки, для величины ошибки в минимуме $\mathbf{S_0}$ получим:

$$P = \frac{1}{\sqrt{2\pi}\,\Phi(r)} \int_0^\infty \Phi\left(-2x\sqrt{m(m+1)}\right) e^{-\frac{1}{2}(x-r)^2} dx, \qquad (14)$$

где $\Phi(z)$ - стандартный интеграл вероятностей [8], а параметр $r$ с точностью до нормировки есть энергия $E(\mathbf{S_0}) = -\frac{1}{2} r N^{3/2}$ в минимуме $\mathbf{S_0}$. От минимума к минимуму величина $r$ изменяется, флуктуируя с небольшим отклонением $(\sim N^{-1})$ вокруг среднего значения $\langle r \rangle \approx 1.37$. Из (14) следует, что с ростом глубины минимума величина ошибки $P$ быстро уменьшается и эффективность алгоритма минимизации увеличивается.

Выражение (14) хорошо согласуется с экспериментом (см. рис.2). Максимальная величина ошибки $P \approx 0.03$ имеет место при $m = 1$. С ростом числа градаций $m$ величина ошибки убывает. Скорость убывания определяется вытекающим из (14) оценочным выражением:

$$P \approx \frac{1}{\pi^2 \sqrt{\pi + 8m(m+1)}}, \qquad (15)$$

справедливым при $m \gg 1$ (при $m \sim 1$ это выражение дает заниженную на 30% величину).

Эффективность минимизации определяется величиной $\delta E = \left(E_0^* - E_0\right)/E_0$, где $E_0^* = E(\mathbf{S_0^*})$ - энергия минимума $\mathbf{S_0^*}$, в который мы спускаемся с использованием «дискретизированного» решающего правила (4), а $E_0 = E(\mathbf{S_0})$ - энергия ближайшего к нему минимума $S_0$, в который мы бы спустились при использовании корректного решающего правила (2). Исходя из (14)-(15) эффективность минимизации можно оценить хорошо совпадающим (см. рис.3) с экспериментом выражением $\delta E \approx 2P$, из которого следует, что разница в энергиях с ростом $m$ убывает как $\delta E \sim m^{-1}$.

6. Обсуждение результатов. Проведенный выше анализ показал, что минимизация функционала $E$ будет достаточно эффективной, если дискретизацию провести в оптимальном виде: задать величину $a_0$ и вектор **b** в виде (9) и оптимизировать величину (11) подбором длины $l_0$ нулевого участка. Тогда величина ошибки $P$ – вероятность рассогласования направлений градиентов исходного $E$ и его «дискретизированного» аналога $\varepsilon$ - будет предельно малой. Действительно, как следует из (13)-(14) величина ошибки даже в простейшем случае $m = 1$ в среднем не превышает значения $P \sim 0.07$ (снижается от значения 0.11 в случайной точке до 0.03 в минимуме). С ростом числа градаций величина ошибки в соответствии с (13) и (15) резко падает. Так, при $m = 16$ средняя величина ошибки не превышает значения $P \sim 0.003$ (по мере спуска ошибка снижается от значения $8*10^{-2}$ в случайной точке до $7*10^{-4}$ в минимуме).

Мы анализировали величину ошибки $P$, поскольку это единственный параметр, который мы можем анализировать в явном виде. Конечная наша цель – показать, что минимумы градиентов исходного функционала $E$ и его «дискретизированного» аналога $\varepsilon$ находятся на небольшом расстоянии $d$ и разница $\delta E$ энергий этих минимумов невелика. Действительно, расстояние между минимумами функционала $E$ и его «дискретизированного» аналога $\varepsilon$ можно оценить простым соотношением $d = NP$, находящемся в хорошем согласии с экспериментом. Зависимость величины $d$ от параметра дискретизации $m$ приведена на рисунке 4. Как видим, расстояние между минимумами достаточно мало: не превышает значения $d = 0.11N$ при $m = 1$ и снижается до значения $d = 0.02N$ при $m = 16$. Зависимость величины $\delta E = 2P$, характеризующей эффективность минимизации, приведена на рисунке 3. Как видим, эффективность «дискретизированного» алгоритма достаточно велика: разница в энергиях составляет всего лишь 7% при $m = 1$ ($\delta E \sim 0.07$) и меньше 0.2% при $m = 16$.

Чтобы понять, насколько введение отрезка с нулевым значением влияет на эффективность минимизации, сравним полученные здесь результаты с результатами работ, использующими стандартный подход: дискретизация матрицы проводится без включения нулевого отрезка ($l_0 = 0$) и элементы бинаризованной матрицы задаются в виде $a_{ij} = sign(A_{ij})$. На рисунке 1 для сравнения приведена пунктирная кривая, построенная по данным эксперимента [5], проведенного при $B_i = 0$ с бинаризованной матрицей. Сравнение показывает, что при $m = 1$ введение нулевого отрезка в 2 раза уменьшает величину ошибки $P$, в 1.5 раза уменьшает расстояние $d$ между минимумами (см.рис.4) и почти в 3 раза уменьшает величину $\delta E$ (см.рис.3). При $m>1$ эти преимущества только увеличиваются.

Еще раз подчеркнем, что описывающие процесс дискретизации выражения (6)-(11) получены в самом общем виде, без привязки к виду распределения матричных элементов $A_{ij}$ и типу выбранного метода дискретизации. Сказанное, прежде всего, относится к параметру $\rho_{min}$, описываемому выражением (11). Этот параметр зависит от типа матрицы $A_{ij}$. Рассчитав этот параметр можно заранее определить верхний предел ошибки $P$ и, тем самым, определить возможность применения процедуры дискретизации. Выше мы показали, что дискретизацию вполне успешно можно применять для матриц, элементы которых подчиняются равномерному распределению. Экспериментальная проверка показала, что дискретизация эффективна в случае матриц хэббовского типа (корреляционные матрицы), матриц модели Изинга и матриц с нормальным распределением. Однако, наверняка существуют матрицы, у которых величина $\rho_{min}$ столь мала (велика величина ошибки), что применить дискретизацию невозможно.

Дискретизация матричных элементов приводит к существенному снижению объема требуемой оперативной памяти и вычислительной сложности алгоритма. Например, при $m=1$ потребность в оперативной памяти и вычислительная сложность алгоритма снижаются в 16 раз. Если эффективность алгоритма, достигнутая при $m=1$ недостаточна, можно перейти к числу градаций $m=16$, уменьшив скорость алгоритма в 2 раза, но существенным образом повысив его эффективность (уменьшая величины $d$ и $\delta E$).

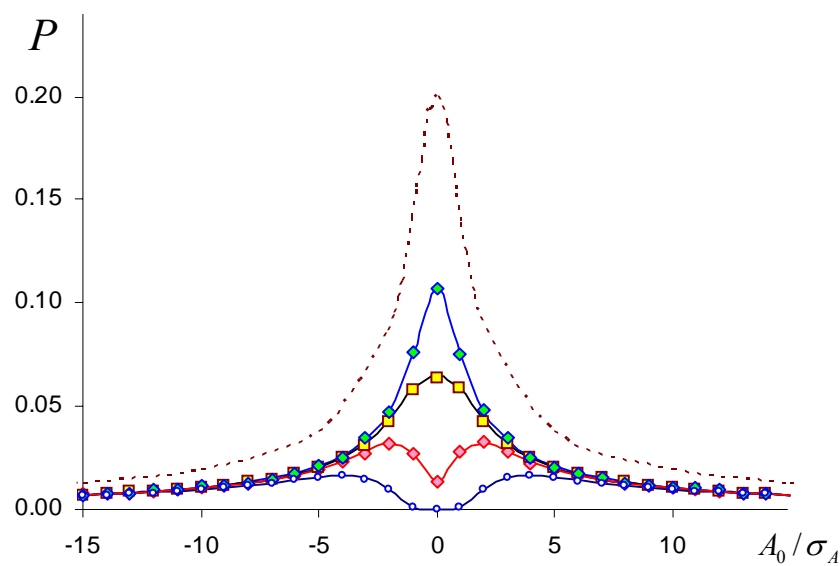

Рис.1  Зависимость ошибки $P$ от величины среднего $A_0$ при различных значениях величины $B_i/\sqrt{N}\sigma_A = 0,1,2,4$. Сплошные кривые – теория, маркеры – эксперимент. Пунктирная кривая - данные эксперимента [7] с бинаризованной матрицей.

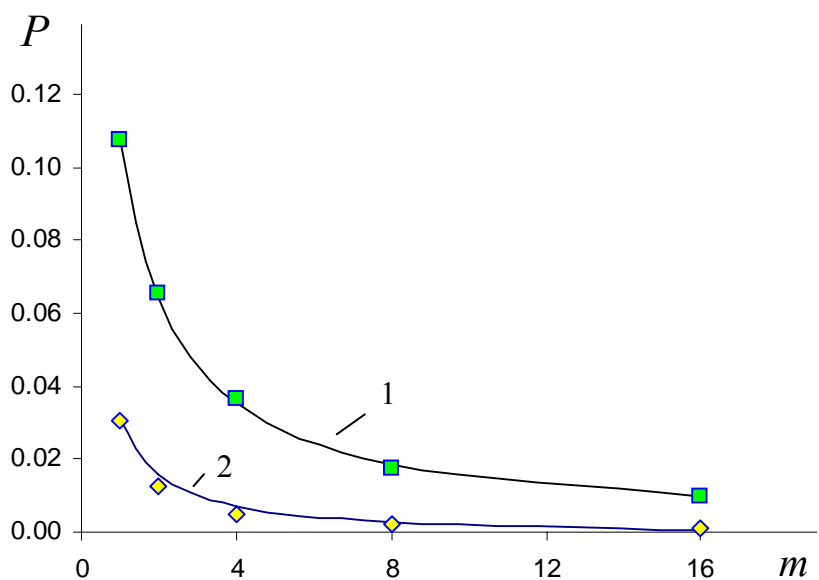

Рис.2  Зависимость величины ошибки от числа градаций m при дискретизации матрицы с равномерным распределением элементов: 1 – ошибка в случайной точке пространства, 2 – ошибка в минимуме. Сплошные кривые – теория, маркеры – эксперимент.

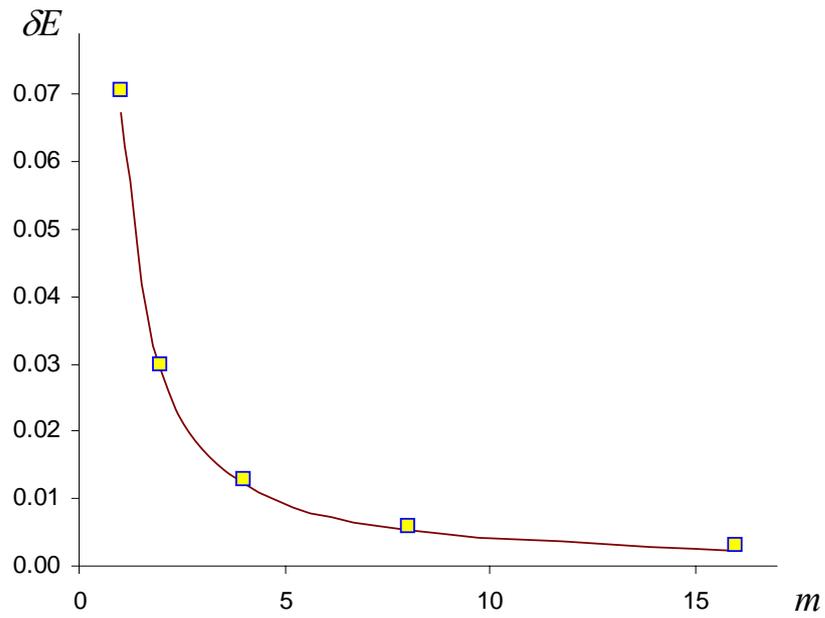

Рис.3 Разница энергий в минимумах исходного и дискретизированного функционалов: сплошная кривая – теория, маркеры – эксперимент.

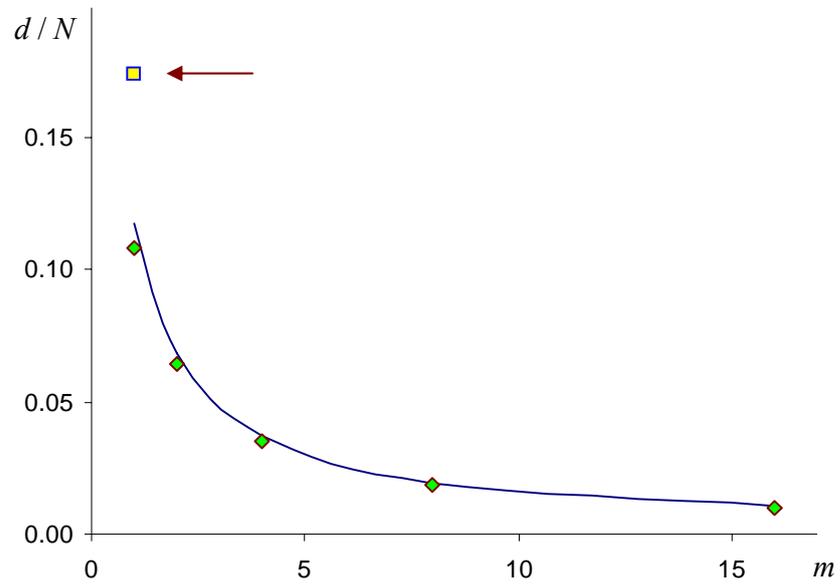

Рис. 4. Расстояние между минимумами исходного и дискретизированного функционалов: сплошная кривая – теория, маркеры – эксперимент. Стрелка указывает на точку $d \approx 0.18N$, взятую для сравнения из экспериментов [7] с бинаризованной матрицей.

Научно исследовательский институт системных исследований РАН (НИИСИ РАН)

117218 Москва, Нахимовский пр-т, 36-1

Крыжановский Борис Владимирович, тел./факс (499)135-13-51 (раб), kryzhanov@mail.ru

Крыжановский Михаил Владимирович, тел. (499)135-78-02 (раб) , iont.niisi@gmail.com

Мальсагов Магомед Юсупович (автор для переписки), тел. (499)135-78-02 (раб), Magomed.malsagov@gmail.com